\documentclass{article}

\ifdefined\pdfsuppressptexinfo
\fi

\usepackage[dblblindworkshop,final]{neurips_2026}

\usepackage[utf8]{inputenc} 
\usepackage[T1]{fontenc}    
\usepackage{url}            
\usepackage{booktabs}       
\usepackage{amsfonts}       
\usepackage{nicefrac}       
\usepackage{microtype}      
\usepackage{xcolor}         
\usepackage{amsmath}
\usepackage{graphicx}
\usepackage{subcaption}
\usepackage{tikz}
\usetikzlibrary{arrows.meta}
\usepackage{wrapfig}
\usepackage{placeins}
\usepackage{float}
\usepackage{arydshln}

\usepackage{xcolor}         
\definecolor{LinkBlue}{HTML}{1F4E79}
\definecolor{CiteGreen}{HTML}{00843D}
\definecolor{UrlBlue}{HTML}{176B87}
  \usepackage[
    colorlinks=true,
    linkcolor=LinkBlue,
    citecolor=CiteGreen,
    urlcolor=UrlBlue
  ]{hyperref}

\usepackage{amsmath, amssymb, amsfonts}
\usepackage{mathtools}
\usepackage{bm}

\newcommand{\R}{\mathbb{R}}

\title{The Optimization Landscape of Learning Compacted Context Models}

\workshoptitle{Continual Learning in the Era of Foundation Models and Embodied Agents}

\author{%
  \textbf{Thomas Villeneuve} \\
  Base Labs \\
  {\footnotesize\texttt{thomas.villeneuve@baseten.co}}
  \And
  \textbf{Alex Sandomirsky} \\
  Base Labs \\
  {\footnotesize\texttt{alex.sandomirsky@baseten.co}}
  \And
  \textbf{Charles O'Neill} \\
  Base Labs \\
  {\footnotesize\texttt{charles.oneill@baseten.co}}
  \AND
  \textbf{Max Kirkby} \\
  Base Labs \\
  {\footnotesize\texttt{max.kirkby@baseten.co}}
  \And
  \textbf{Michael Psenka} \\
  Base Labs \\
  {\footnotesize\texttt{psenka@baseten.co}}
}

\begin{document}

\maketitle
\suppressfloats[t]

\begin{abstract}
Many works approach continual learning through the lens of infinite context windows. As an agent puts more observation into context (concretely the KV cache), compacting said context is akin to direct memory manipulation, without affecting the base model's weights. Many works pose KV compaction as an optimization problem: learn a smaller set of KV vectors that matches the behavior of the full KV cache. While this preserves base model behavior, optimizing through a frozen base model results in a highly nontrivial optimization problem with a brittle and flat loss landscape. In this paper, we characterize what makes these optimization problems difficult and demonstrate that a heavily simplified Perceiver-based architecture not only matches performance of a full Perceiver transformer in continuous context compaction, but outperforms baselines on compaction utility. Results are presented on MCQ tasks across Finance, Legal, Gutenberg, and Code.
\end{abstract}

\section{Introduction}

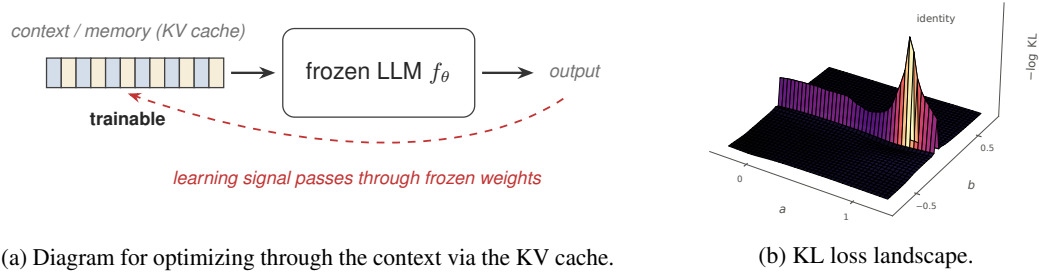
\begin{figure}[t]
    \centering
    \begin{subfigure}[c]{0.58\linewidth}
        \centering
        \resizebox{\linewidth}{!}{
\begin{tikzpicture}[
    x=1mm, y=1mm,
    font=\sffamily\scriptsize,
    ink/.style={draw=black!75, line width=0.5pt},
    dim/.style={text=black!55, font=\sffamily\itshape\scriptsize},
    flow/.style={-{Stealth[length=2.2mm]}, line width=0.9pt, draw=black!80},
    grad/.style={-{Stealth[length=1.8mm]}, line width=0.7pt, draw=red!70!black!80, dashed},
]
\definecolor{kfill}{RGB}{212,222,235}
\definecolor{vfill}{RGB}{247,238,218}

\foreach \i in {0,...,5} {
  \draw[ink, fill=kfill] (\i*4,14) rectangle ++(2,4);
  \draw[ink, fill=vfill] (\i*4+2,14) rectangle ++(2,4);
}
\node[dim] at (11,21.5) {context / memory (KV cache)};
\node[text=black!85, font=\sffamily\bfseries\scriptsize] at (11,9.8)
    {trainable};

\draw[ink, fill=white, rounded corners=1.6mm] (32,10) rectangle (58,22);
\node[text=black!85, font=\sffamily\small, align=center] at (45,16)
    {frozen LLM $f_\theta$};

\draw[flow] (25,16) -- (31,16);
\draw[flow] (59,16) -- (66,16);
\node[dim, anchor=west] at (66.8,16) {output};

\draw[grad] (70,12.5) .. controls (58,4.5) and (36,4.5) ..
      (11,13.2);
\node[dim, text=red!70!black!80] at (42,1.6)
    {learning signal passes through frozen weights};
\end{tikzpicture}}
        \vspace{0.19em}
        \caption{Diagram for optimizing through the context via the KV cache.}
        \label{fig:overview-setting}
    \end{subfigure}\hfill
    \begin{subfigure}[c]{0.35\linewidth}
        \vspace{-4em}
        \centering
        \includegraphics[width=\linewidth]{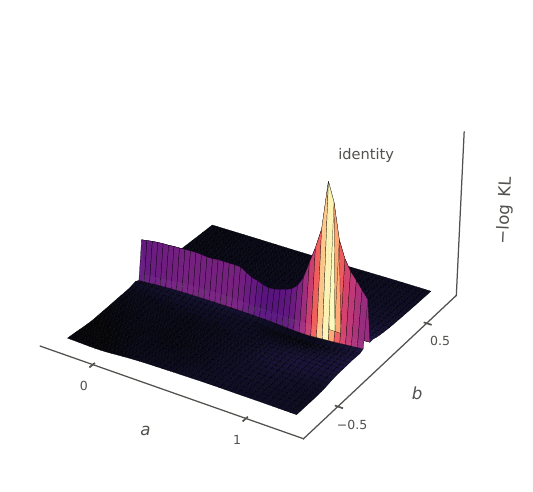}
        \vspace{-2em}
        \caption{KL loss landscape.}
        \label{fig:overview-landscape}
    \end{subfigure}
    \vspace{0.2em}
    \caption{Optimizing directly through context is appealing in keeping as much of the main model behavior as possible, but the loss landscape can be quite sharp. The loss landscape is given by the 2-parameter family in eq.~\eqref{eq:2param_matrix}, applied to all heads and layers instead of one, thus being much sharper than Figure~\ref{fig:kvmix2x2-landscape-broad}.}
    \vspace{1em}
    \label{fig:overview}
\end{figure}

A key component of any continually-deployed model, whether text-based or robotic, is its memory management; what to keep, what can be forgotten, and what can be compressed down. Most language models today are at least in some part attention-based: as tokens are observed and the context grows (in the form of the KV cache), computation time scales quadratically, and observed tokens are less attended to. As such, rather than designing new architectures that bound memory or linearly scale with context, a growing line of work revolves around compacting the KV cache: either by evicting entries \citep{zhang2023h2o,
li2024snapkv}, by optimizing a cache per document
\citep{eyuboglu2025cartridges, zweiger2026fast}, or by training a model
that writes compact entries \citep{chari2025kvdistill, o2026still}. We
refer to these kinds of models as \emph{compacted context models}.

However, this optimization can often be difficult: per-document methods spend minutes of
optimization per document, and optimizing through a fixed deep model introduces a highly nonlinear, highly non-convex structure. Without a deeper understanding of where and how these optimization difficulties arise, many of the same mistakes are silently made in parallel.

This paper provides a deeper characterization of the difficulty in optimizing through KV caches, starting from the smallest setting that
exhibits the difficulty. A compression-free linear map already showcases difficulty in the optimization: near-identity maps can still fall into bad local minima. Measuring the loss surface directly shows
why: the global solution sits in a needle-sharp basin surrounded by a wide,
flat plateau (Figure~\ref{fig:overview}b).

Each of the loss landscape characteristics point to a specific potential fix that has been explored in some form in the literature. For the flat landscape, one could hope that a loss more densely supervised over the model could unflatten the landscape. This could be through the base model's forward pass in regulating attention passes through the full network the same way, or through more generated self-study examples on the input context. For the sharp basin, one could hope that relaxing the base model weights could make the overall optimization smoother and more tractable. However, we find that attempted fixes to the KV landscape quickly fail once compressions ratios increase, or compactions are iterated on each other (an inevitability for continually compacted memory models).

\begin{figure}[t]
    \centering
    \resizebox{\linewidth}{!}{
  \begin{tikzpicture}[
      x=0.7mm, y=0.5mm,
      every node/.append style={scale=0.55, transform shape},
      font=\sffamily\scriptsize,
      ink/.style={draw=black!75, line width=0.3pt},
      lbl/.style={text=black!75},
      dim/.style={text=black!55, font=\sffamily\itshape\scriptsize},
      flow/.style={-{Stealth[length=2.0mm]}, line width=0.9pt, draw=black!80},
      proj/.style={-{Stealth[length=1.5mm]}, line width=0.7pt, draw=black!65},
      match/.style={dash pattern=on 1pt off 1pt, line width=0.6pt,
                    draw=orange!60!black!75},
      read/.style={-{Stealth[length=1.4mm]}, line width=0.5pt,
                   draw=orange!60!black!75},
  ]

  \definecolor{kfill}{RGB}{212,222,235}
  \definecolor{vfill}{RGB}{247,238,218}
  \definecolor{zfill}{RGB}{248,213,190}
  \definecolor{klabel}{RGB}{110,138,175}
  \definecolor{vlabel}{RGB}{186,124,84}
  \definecolor{sheet}{RGB}{250,249,247}
  \definecolor{attn}{RGB}{200,120,70}
  \definecolor{kproj}{RGB}{190,224,223}
  \definecolor{vproj}{RGB}{223,232,189}

  \newcommand{\plaingrid}[6]{
    \pgfmathsetmacro{\w}{#4*#5}
    \pgfmathsetmacro{\h}{#3*#5}
    \draw[ink, fill=#6] (#1,#2) rectangle ++(\w,\h);
    \foreach \r in {1,...,\numexpr#3-1\relax}
      \draw[black!40, line width=0.3pt] (#1,#2+\r*#5) -- ++(\w,0);
    \foreach \c in {1,...,\numexpr#4-1\relax}
      \draw[black!40, line width=0.3pt] (#1+\c*#5,#2) -- ++(0,\h);
  }

  \newcommand{\sheetgrid}[6]{
    \pgfmathsetmacro{\w}{#4*#5}
    \pgfmathsetmacro{\h}{#3*#5}
    \draw[ink, fill=sheet, rounded corners=0.6mm]
      (#1+2.4,#2-2.4) rectangle ++(\w,\h);
    \draw[ink, fill=sheet, rounded corners=0.6mm]
      (#1+1.2,#2-1.2) rectangle ++(\w,\h);
    \plaingrid{#1}{#2}{#3}{#4}{#5}{#6}
  }

  \draw[ink, fill=sheet, rounded corners=0.6mm]
    (4.4,5.6) rectangle ++(12.8,28.8);
  \draw[ink, fill=sheet, rounded corners=0.6mm]
    (3.2,6.8) rectangle ++(12.8,28.8);
  \plaingrid{2}{8}{9}{2}{3.2}{kfill}
  \plaingrid{8.4}{8}{9}{2}{3.2}{vfill}
  \draw[ink] (2,8) rectangle ++(12.8,28.8);

  \node[text=klabel, font=\sffamily\itshape\small]
    at (5.2,39.5) {K};
  \node[text=vlabel, font=\sffamily\itshape\small]
    at (11.6,39.5) {V};

  \draw[black!45, line width=0.4pt] (-0.5,8) -- (-0.5,36.8);
  \node[dim, anchor=east] at (-1.5,22.4) {T};

  \plaingrid{32}{27}{6}{2}{3.2}{kproj}
  \node[text=klabel, font=\sffamily\itshape\small]
    at (35.2,48.4) {$\tilde{K}$};

  \plaingrid{32}{-3.6}{7}{2}{3.2}{vproj}
  \node[text=vlabel, font=\sffamily\itshape\small]
    at (35.2,21.2) {$\tilde{V}$};

  \draw[proj] (17.6,28.5) -- (31.2,34);
  \node[dim] at (23.6,34.2) {$W_K$};

  \draw[proj] (17.6,16.3) -- (31.2,10.8);
  \node[dim] at (23.6,10.6) {$W_V$};

  \node[text=zfill!40!black, font=\sffamily\bfseries\small]
    at (63.8,27.2) {Z};

  \foreach \i in {0,1,2,3}
    \draw[ink, fill=zfill] (52+\i*4,20.6) rectangle ++(3.6,3.6);

  \node[lbl] at (70.2,22.4) {$\cdots$};
  \draw[ink, fill=zfill] (72,20.6) rectangle ++(3.6,3.6);
  \node[dim] at (63.8,16.9) {$t$ trainable slots};

  \draw[match] (53.8,24.4) -- (38.4,29);
  \draw[match] (57.8,24.4) -- (38.4,32.5);

  \draw[read] (38.4,15.2) -- (53.2,20.2);
  \draw[read] (38.4,11.8) -- (57.4,20.2);

  \plaingrid{96}{26.4}{3}{2}{3.2}{kfill}
  \node[text=klabel, font=\sffamily\itshape\small]
    at (99.2,38.2) {$C_k$};

  \plaingrid{96}{8.8}{3}{2}{3.2}{vfill}
  \node[text=vlabel, font=\sffamily\itshape\small]
    at (99.2,6) {$C_v$};

  \node[dim] at (99.2,2.6) {$t \ll T$};

  \draw[flow] (77,22.4) -- (94.4,22.4);
  \node[dim] at (85.6,25) {$W_O$};

  \draw[
    {Stealth[length=0.9mm]}-{Stealth[length=0.9mm]},
    line width=0.5pt,
    draw=black!65
  ] (99.2,19.1) -- (99.2,25.7);

  \end{tikzpicture}}
    \caption{The simplified Fast Diamond writer. Architecture details can be found in Section~\ref{sec:fast-diamond}.}
    \label{fig:fastdiamond-diagram}
\end{figure}
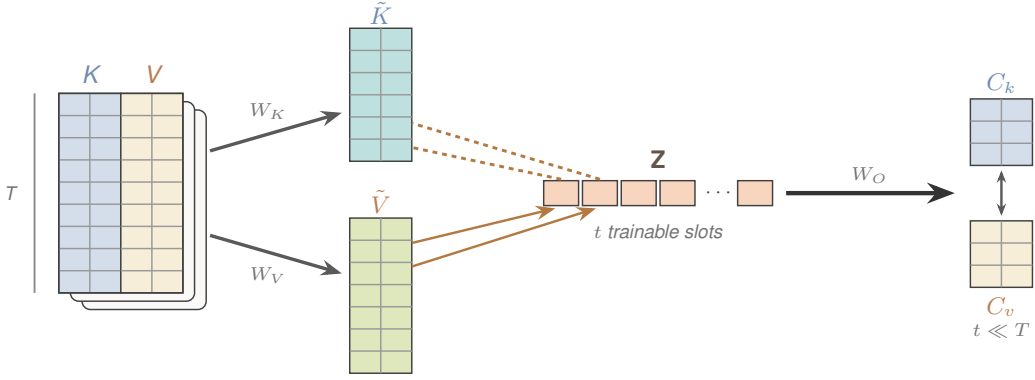

We then find that the most performant compacted context model is a heavily simplified, single-layer single head attention layer, depicted in Figure~\ref{fig:fastdiamond-diagram} and detailed in Section~\ref{sec:fast-diamond}. Our KV compactor beats all compared baselines, both in terms of MCQ accuracy and compaction speed.

In summary, our contributions are:
\begin{enumerate}
    \item A set of synthetic experiments to visualize and characterize the optimization landscape of KV space optimization,
    \item An empirical account of why common fixes, including proxy losses,
    more data, and weight relaxation, do not reliably help, and
    \item A minimal cache writer, which we call Fast Diamond, that matches or beats
    more expressive writers and standard baselines under matched MCQ
    evaluation.
\end{enumerate}


\section{Problem formulation}
\label{sec:problem_formulation}


We borrow notation from~\citep{o2026still}. Consider a pretrained autoregressive transformer $f_\theta$, kept frozen
throughout, with $L$ layers, $H$ KV-heads per layer, head dimension $d$,
and rotary position embeddings
\citep{su2023roformerenhancedtransformerrotary}. A prefix of
$T$ tokens leaves behind a per-layer KV cache: at layer $\ell$ and head
$h$, we notate cached keys and values $K^{(\ell,h)}, V^{(\ell,h)} \in
\mathbb{R}^{T \times d}$. Notate $f_\theta(\;\cdot\; | K, V)$ to be the induced function of $f_\theta$ with the full caches $K, V$ in context. We will consider per-layer functions over cached keys and values:
\begin{equation}
g_\phi^{(\ell)} :
    (K^{(\ell,h)}, V^{(\ell,h)})_{h=1}^{H}
    \mapsto
    (C_k^{(\ell,h)}, C_v^{(\ell,h)})_{h=1}^{H},
\end{equation}
with $C_k^{(\ell,h)}, C_v^{(\ell,h)} \in \mathbb{R}^{t \times d}$,
and $t \ll T$. Denote $g_\phi(K, V)$ as the induced function over the full caches $K, V$.

Our objective is to learn $g_\phi^{(\ell)}$ such that $f_\theta(\;\cdot\; | g_\phi(K, V))$ matches the performance of $f_\theta(\;\cdot\; | K, V)$ as closely as possible, given that $t \ll T$. Rather than trying to fully reproduce all base behavior, our main optimization objectives throughout the paper are KL divergences over the answers of various document-based questions:
\begin{equation}\label{eq:loss_main}
    \mathcal{L}(\phi) = \mathbb{E}_{x \sim \mathcal{D},\, (q, a) \sim Q(x)} \; \mathrm{KL}\Big(f_\theta(a \mid q, K(x), V(x)) \,\|\, f_\theta(a\mid q, g_\phi(K(x), V(x)))\Big).
\end{equation}
We consider the output space of $g_\phi$ to be the full embedding space of $k, v\in \R^d$, rather than the discrete space resulting from forward passes on the finite token dictionary. We then say ``KV space'' to mean the full embedding Euclidean space $\R^d \times \R^d$.

Experiments used eight-GPU NVIDIA H100 80\,GB nodes or eight-GPU
NVIDIA H200 141\,GB nodes; Appendix~\ref{app:compute-resources} gives the
allocation in more detail.

\section{Simplified model experiments: optimizing linear maps in KV space}
\label{sec:toy-landscape}
For our simplest setting, we consider non-compressive bilinear maps over KV space; that is, we optimize a matrix
$W\in\R^{256\times256}$ that acts on the concatenation of $k$ and $v$:
\begin{equation}
    g_\eta(k, v)
    = W \begin{pmatrix}
        k \\ v
    \end{pmatrix}
    =
    \begin{pmatrix}
        W_{kk} & W_{kv} \\ W_{vk} & W_{vv}
    \end{pmatrix}\begin{pmatrix}
        k \\ v
    \end{pmatrix}.
\end{equation}
where each $k,v\in\R^{128}$. Everything else is the same as the full problem formulation in Section~\ref{sec:problem_formulation}.

\begin{figure}[t]
    \centering
    \begin{subfigure}[t]{0.44\linewidth}
        \centering
        \includegraphics[width=0.92\linewidth]{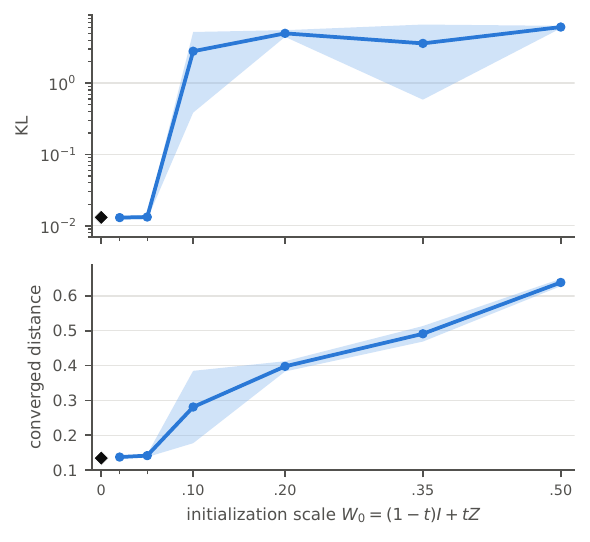}
        \caption{\emph{The loss landscape is sharp around the global optimum.} Near-identity iterates sharply become unstable as we gradually increase init norm.}
        \label{fig:init-variance-sweep}
    \end{subfigure}\hfill
    \begin{subfigure}[t]{0.46\linewidth}
        \centering
        \includegraphics[width=0.95\linewidth]{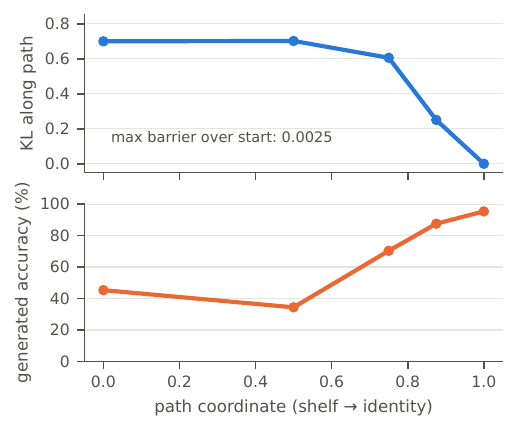}
        \caption{\emph{The global minimum is not in an isolated basin}, there exist constructed paths with minimal energy increase that reach the global optimum.}
        \label{fig:shelf-identity-path}
    \end{subfigure}
    \caption{Reachability of the global optimum for the linear KV compression model. Diamonds on the LHS indicate a control where we run optimization starting from the global optimum $I$, indicating the amount of loss/distance resulting from precision error. The RHS waypoints are constructed by first linearly interpolating between initialization and $I$, then iteratively (a) optimizing the max loss among waypoints, and (b) redistribute for constant arc length along the interpolated piecewise-linear curve.}
    \label{fig:toy-landscape}
\end{figure}

\begin{wrapfigure}[19]{r}{0.45\linewidth}
    \centering
    \vspace{-1em}
    \includegraphics[width=\linewidth]{
        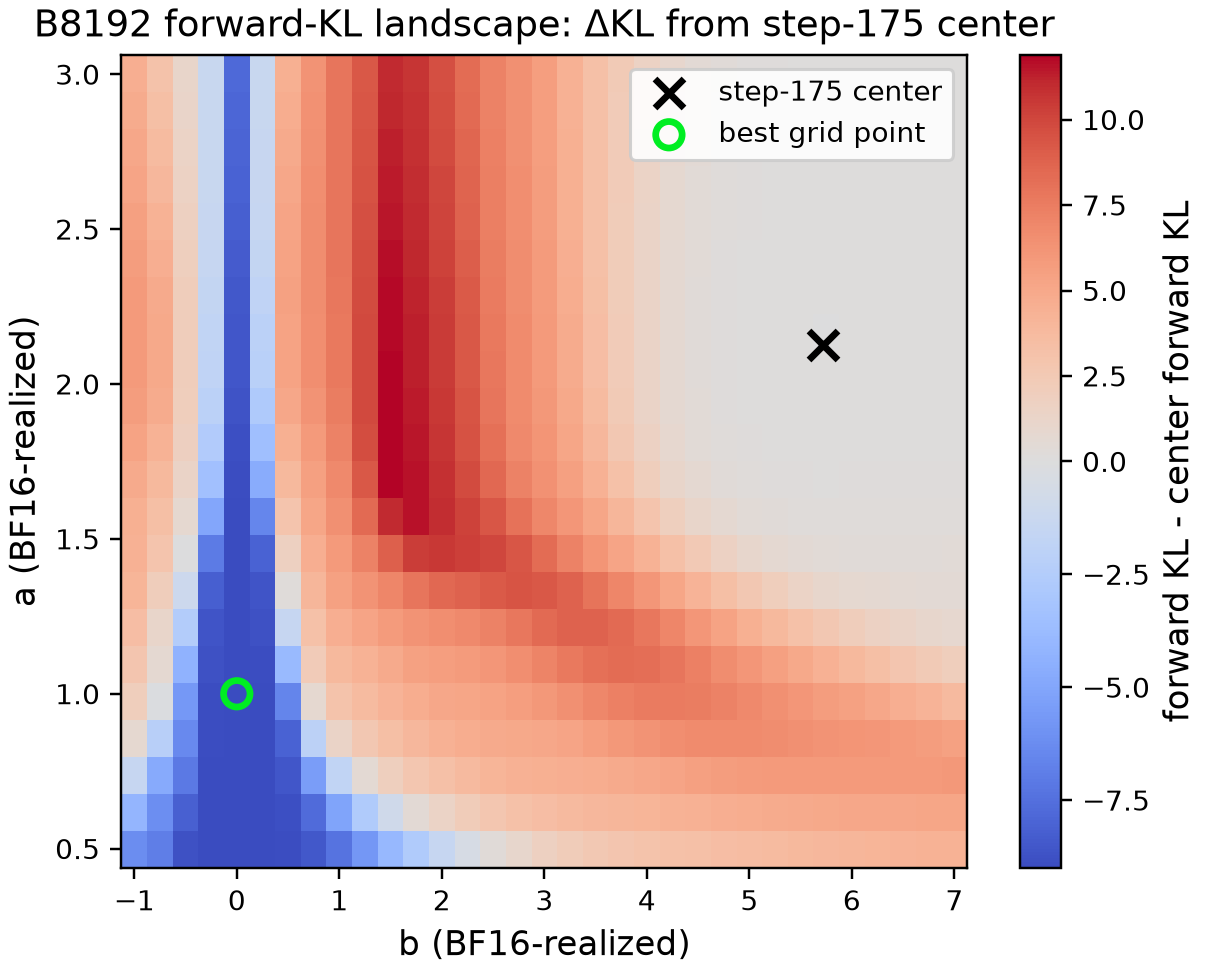
    }
    \caption{Loss landscape of KV compression with respect to the 2-parameter family given in eq.~\eqref{eq:2param_matrix}, single head/layer.}
    \label{fig:kvmix2x2-landscape-broad}
\end{wrapfigure}

While the landscape seems stable very close to the
identity, there are many wide local minima further away from the global
minimum. Some of the landscape seems due to the base model geometry, and
some seems due to inner cancellations; for example, gradients between data
points (Figure~\ref{fig:init-variance-sweep}).


Instead of immediately jumping into the full landscape, we consider 2-dimensional reductions to be able to fully visualize the loss landscape structure, starting with the following:
\begin{equation}\label{eq:2param_matrix}
    W(a,b)=
    \begin{bmatrix}
        a & b \\
        b & a
    \end{bmatrix}
    \otimes I_{128},
\end{equation}
i.e. symmetric and isotropic within all four K/V blocks, so the identity solution here is $(a,b)=(1,0)$.
Applying this matrix only to a single head's KV cache in the first layer is
already enough to observe a pathological loss landscape.
Figure~\ref{fig:kvmix2x2-training} shows the loss curve obtained from training
this setup with an initialization of $(a_0,b_0)=(3,3)$.

With only two parameters, we can visualize the historical loss slice in
Figure~\ref{fig:kvmix2x2-landscape}, which immediately displays two important characteristics: \textbf{(a)} the region of the global optimum is narrow and sharp, and \textbf{(b)} the region of the bad local optimum is wide and flat.

This then gives a more visual picture to what we observe in the full matrix optimization case for Figure~\ref{fig:toy-landscape}: while gradient and momentum-based optimization plateaus quite early, the reason is for a strikingly flat landscape rather than a sharp/brittle one, and while the path into the global minimum is sharp, there do exist connecting paths with low energy barriers.

\section{Attempts to improve the landscape}
\label{sec:landscape-fixes}

We now go through common fixes that have been explored in the literature, each tied to a specific geometric aspect of the loss landscape depicted in Figure~\ref{fig:kvmix2x2-landscape-broad}.

\subsection{A denser loss signal}
\label{sec:fix-denser}
Flatness of the bad local minima indicates we may need a loss that is more densely supervised than eq. \eqref{eq:loss_main}. One way to supervise further is to have the query tokens pass through the full model in the same way, e.g. by supervising an $L^2$ norm over all layers:
\begin{equation}
    \mathcal{L}_\text{AM}(\phi) = \sum_{i=1}^L\|f_\theta^i(a \mid q, K, V) - f_\theta^i(a\mid q, g_\phi(K(x), V(x)))\|_2^2,
\end{equation}
where $f_\theta^i(a | \ldots)$ is the internal activation at layer $i$ for input token $a$. This type of layer-dense squared distance loss has been used on a number of KV compression algorithms to provide a denser signal~\citep{onerec2026ksa,gao-etal-2026-sada,zweiger2026fast,jiang2026kvsculpt}.

\begin{wrapfigure}[18]{r}{0.46\textwidth}
    \centering
    \includegraphics[width=0.9\linewidth]{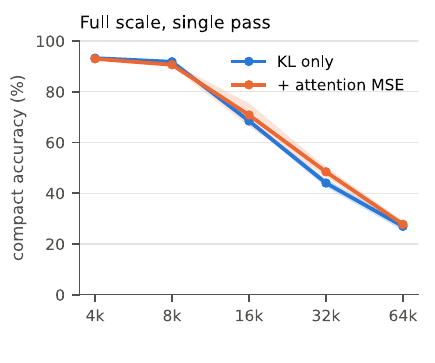}
    \caption{
    Attention MSE does not help KV optimization in compressive regimes (8k$\to$1k tokens). Both decay at similar rates.}
    \label{fig:attnmse-counterexample}
\end{wrapfigure}

For linear map experiments, this greatly improves the optimization landscape, and the global optimum at the identity becomes a much more attractive basin (see e.g. Figure~\ref{fig:app-single-matrix} in the appendix for results). However, these benefits do not carry over to compressive settings where $g_\phi$ outputs fewer tokens. We observe that the internal squared norm can be sensitive with respect to the target KL loss in eq.~\eqref{eq:loss_main}, see e.g. Figure~\ref{fig:app-attnmse-counterexample-toy} in the appendix.
And in the token compressive setting (8k$\to$1k), adding this internal squared norm loss does not meaningfully help the trained performance
(Figure~\ref{fig:attnmse-counterexample}).




The failures above suggest three seemingly attractive fixes: give the
optimizer a denser signal, add self-study queries or relax the frozen weights.
None of them reliably worked, and the ways they fail are informative.


\subsection{Adding self-study queries}
\label{sec:fix-selfstudy}

While we train on a dataset of MCQ questions, it may be helpful to further optimize on \emph{self-study queries}, i.e. using the base model to generate questions about the document, then matching answer-token KL between the full and compressed contexts. This process is at the core for a number of KV compression algorithms, notable Cartridges~\citep{eyuboglu2025cartridges}. In our linear setting further self-study queries beyond our training dataset do not notably help to find the global optimum.

\subsection{Relaxing the frozen weights}
\label{sec:fix-relaxing}
A reasonable hypothesis for the sharpness of the landscape is optimization through a deep frozen model. A natural way to amend this sharpness is to allow the weights to train, but full training will now heavily overparameterize this optimization problem, potentially leading to many more non-generalizing global minima. We then hope that a small amount of adaptation will be enough to smooth the optimization while keeping as much of the base capability as possible. Other works have explored both lower-rank weight updates~\citep{chevalier2023adapting} or full model weight updates ~\citep{nawrot2024dynamic,li2026lclm}. Unfortunately though, smaller adaptation does not notably improve converged results, and notable gains only appear once full
projections are trainable. Still, even fully training weights does not recover the full performance gap between the compressed and full contexts.

\emph{Temporary weight relaxation}, allowing the full weights to train but use a penalty function to smoothly return them to base weights, gives an interesting alternative that theoretically smooths the loss landscape while allowing converged points to remain at the base model weights. We detail the objective and schedule in Appendix~\ref{app:w-ablate}. 
In toy-still experiments, temporary weight relaxation on only $W_O$ is sufficient to
recover the identity, but in the iterative compaction context it provides little benefit beyond keeping weights frozen.

\begin{figure}[t]
    \centering
    \begin{subfigure}[t]{0.48\linewidth}
        \centering
        \includegraphics[width=\linewidth]{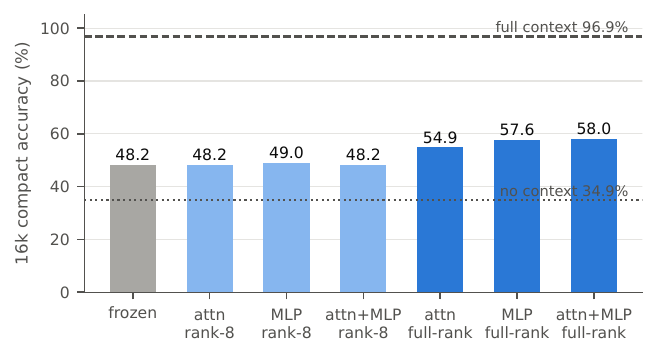}
        \caption{Various scales of base weight relaxation.}
        \label{fig:weight-adaptation}
    \end{subfigure}
    \hfill
    \begin{subfigure}[t]{0.48\linewidth}
        \centering
        \includegraphics[width=0.7\linewidth]{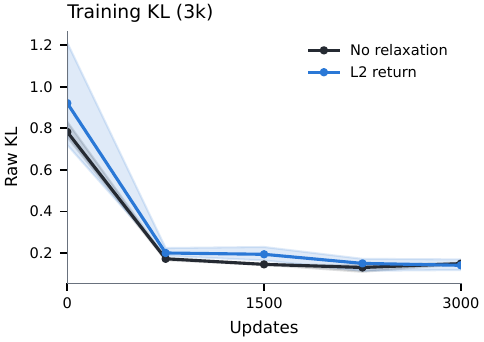}
        \caption{Optimization trajectory of frozen weights vs. temporary weight relaxation.}
        \label{fig:weight-relaxation-review}
    \end{subfigure}
    \caption{Limitations of relaxing the base model weights. The LHS is run in the iterative compaction setting, where we find benefits are not visible until large parts of the model are trainable, even with which the MCQ performance is far from saturated. With attempted temporary weight relaxation in the RHS, in the compressive setting (1k$\to$128), loss similarly flatlines around the same position.}
    \label{fig:weight-relaxation-training-kl}
\end{figure}

\paragraph{Recurrent compaction.}

Armed with a clearer view of the loss landscape and potential corrections, we attempt the single linear map experiment in the recurrent setting.

Concretely, we apply the linear map on itself $n$ times prior to applying the map to the raw KV cache for KV reconstruction and loss calculation:
\begin{equation}
    g_\eta(k, v,n)
    = W^n \begin{pmatrix}
        k \\ v
    \end{pmatrix}
    =
    \begin{pmatrix}
        W_{kk} & W_{kv} \\ W_{vk} & W_{vv}
    \end{pmatrix}^n\begin{pmatrix}
        k \\ v
    \end{pmatrix}.
\end{equation}
where each $k,v\in\R^{128}$. 

Attention MSE does not provide much uplift in the recurrent setting (identity solution remains unfound), however weight training sees real uplift, increasing utility from 35\% to 50-60\% (Appendix~\ref{app:w-ablate}). Both results agree with later experiments in the recurrent Fast Diamond fixed budget setting.

We also find lower KL under recurrence with partial identity constraints on the key map (Appendix~\ref{app:a-ablate}). Fixing 20\% of keys to identity reduces KL from 2.01 to 1.23 after 31 applications, despite training with only seven applications. This suggests that reducing the freedom of the transformation can make recurrence easier to learn, motivating the freezing of more of the compactor, as well as reducing the expressivity of the compactor altogether.




\section{A simplified KV compaction architecture}
\label{sec:fast-diamond}

Given the difficulty of the base optimization problem, we explore the performance ceiling of considering a heavily simplified KV compression architecture. Our starting point is a two-block Perceiver with latent self-attention, learned query projections, cross-attention residuals and trainable output readouts. Removing the second block and self-attention and fixing the output readouts improves recurrent MCQ accuracy from 64.5\% to 69.0\%. Using the latent queries directly and removing the cross-attention residual further improves accuracy to 82.5\% (Appendix~\ref{app:a-ablate}). This leaves a single-head attention layer over joint KV space, with trainable internal projections, which we call Fast Diamond. A number of these are similar modifications made in~\citep{o2026still}.

Formally, per layer the writer reads the cache as
$X = [K; V] \in \mathbb{R}^{T \times 2d}$, with keys un-rotated into a
position-free frame, and maintains $t$ latent queries
$Z \in \mathbb{R}^{t \times 2d}$. One cross-attention block computes
\begin{equation}
    Z' = \mathrm{CrossAttn}(Z, X),
\end{equation}with a single head: $\tilde q = Z$, $\tilde k = X W_K$, $\tilde v = X W_V$;
queries and keys are L2-normalized, RoPE is applied to both, and the
cross-attention output is projected back to the latent space:
\begin{equation}
    Z' = \mathrm{softmax}(d_\ell\, \tilde q \tilde k^\top)\, \tilde v \cdot W_O.
\end{equation}
The three trainable projections $W_K$, $W_V$, and $W_O$ are the only linear
maps the writer learns. The final latent state is read out to compact keys and values by frozen
output heads:
\begin{equation}
    C_k = Z' W_{\mathrm{key}}, \qquad C_v = Z' W_{\mathrm{value}},
\end{equation}
with compact keys re-rotated to cache positions. We initialize $W_{\mathrm{key}}$ and $W_{\mathrm{value}}$ as truncated identity readouts (detailed in Appendix~\ref{app:methods}), but leave them trainable. We denote this the \emph{Fast Diamond} architecture.

\section{Results}
\label{sec:full-method}

We now evaluate our simplified architecture, denoted Fast Diamond, against a suite of baseline methods over multiple-choice questions (MCQ) from on the compressed cache.

\subsection{Data and evaluation}
  Akin to \citep{o2026still}, Fast Diamond and compatible baseline methods are trained on a four-domain extractive MCQ corpus containing Financial filings,
  Project Gutenberg books, Legal documents, and Code\footnote{Code and data to be released for camera-ready.}. Each example contains a document prefix,
  an MCQ prompt, and an answer target. Questions are generated from document spans, verified
  using the frozen base model with context, and filtered to remove questions that it can answer
  without context. Training and evaluation documents are separated at the document level.
  Appendix~\ref{app:methods} gives the data sources, split construction, and per-domain counts.

  The matched single-pass comparison uses a separate held-out set of 256 medical transcripts (outside of the training domains)
  and 758 extractive MCQs. Each transcript contains 8,192 tokens, which each method compresses
  to 1,024 latents.

\subsection{Baseline}
\label{sec:matched-baselines}

We evaluate every method on the same 256 documents and 758 questions with Qwen3-4B using greedy decode. Full-context and no-context inference 
provide the upper and lower controls.

The comparison includes four classes of retained context:  Cartridges~\citep{eyuboglu2025cartridges} learns a separate cache for each document, H2O ~\citep{zhang2023h2o} and Highest Attention Keys matching (HAK) ~\citep{zweiger2026fast} perform training-free document-specific selection, $\delta$-mem ~\citep{lei2026deltamem} maintains a small recurrent state alongside a recent raw-KV window, and Stills amortizes the learning of cache compaction in weights.


We report normalized utility, called \emph{utilization} or \emph{utility} elsewhere in our
experiments  as
\begin{equation}
    U(m)
    =
    100
    \frac{A_m-A_{\mathrm{no\ context}}}
         {A_{\mathrm{full\ context}}-A_{\mathrm{no\ context}}},
    \label{eq:normalized-utility}
\end{equation}
where $A_m$ is the MCQ accuracy of method $m$. Full context and no context have utility
$100\%$ and $0\%$.


\begin{table*}
\centering
\caption{Main results on MCQ compaction. H2O, HAK, and Still measure online compaction time, LCLM reports encoder-plus-adapter compaction time, and Cartridges reports offline synthesis plus per-document optimization time.}
\label{tab:matched-baseline-results}
\small
\resizebox{\textwidth}{!}{%
\begin{tabular}{lrrrrr}
\toprule
Method &
Accuracy &
Normalized utility &
Cache / compaction &
Generation \\
&
{(\%)} &
{(\%)} &
{(s/document)} &
{(s/question)} \\
\midrule
Full-context control
    &
    98.29 & 100.00 & 0.00 & 8.44 \\
    \hdashline
    \rule{0pt}{2ex}%
Fast Diamond
    &
    \textbf{92.48} & \textbf{93.20} & \textbf{0.05} & 10.89 \\
LCLM mean-pooling 8$\times$
    &
    90.11 & 90.42 & 0.07 & \textbf{1.45} \\
HAK
    &
    83.38 & 82.53 & 56.26 & 16.18 \\
Cartridges
    &
    80.61 & 79.20 & 947.19 & 9.65 \\
H2O
    &
    74.01 & 71.56 & 0.43 & 10.61 \\
$\delta$-mem + recent-1024 KV
    &
    25.59 & 14.84 & 0.42 & 3.11 \\
Recent-1024 KV only
    &
    22.43 & 11.13 & 0.24 & 1.84 \\
    \hdashline
    \rule{0pt}{2ex}%
No-context control
    &
    12.93 & 0.00 & 0.00 & 10.13 \\
\bottomrule
\end{tabular}%
}

\vspace{2pt}
\end{table*}

\subsection{Fixed-budget comparison}
\label{sec:fixed-budget-streaming}

\begin{wrapfigure}[21]{r}{0.53\linewidth}
    \centering
    \includegraphics[width=0.5\textwidth]{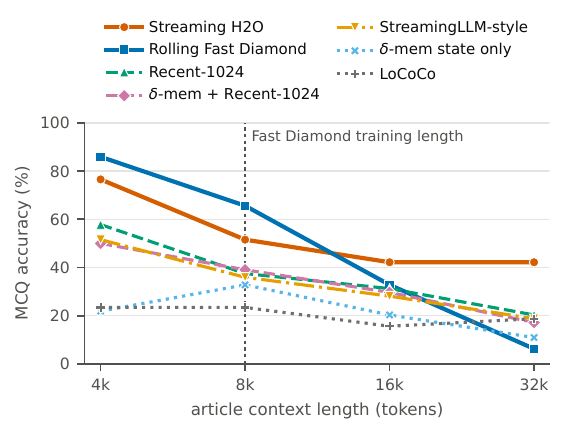}
    \caption{Iterative compaction results. Compactions occur every 1k tokens. While our simplified architecture Fast Diamond has the highest-performing compressed KV cache at earlier iterations, it drops off into later compactions (16k and 32k) in favor of Streaming H20.
    }
    \label{fig:fixed-budget-context-curves}
\end{wrapfigure}

For the fixed-budget evaluation, every method has a persistent document-memory budget of 1k (1,024) positions, while the question
  and generated answer remain outside this budget. We report raw MCQ performance in Figure~\ref{fig:fixed-budget-context-curves}.

LoCoCo~\citep{cai2024lococo} and $\delta$-mem are techniques that do continuous compaction, and so we benchmark them techniques against Still along with custom variants of StreamingLLM ~\citep{xiao2024efficient} and H2O to draw the frontier of fixed budget techniques in continuous compaction methods. 



Rolling Fast Diamond (fixed budget) applies the Section~\ref{sec:fast-diamond} writer recurrently: taking as input two chunks (whether compacted latents or raw KV) and outputs compacted latents:

$K_0 = C_k$, $K_1 = K_{0:c}$ $V_0 = C_v$, $V_1 = V_{0:c}$ where $c$ denotes the chunk size and 0 indexes the oldest non compacted token. $X = [K_0, K_1; V_0, V_1] \in \mathbb{R}^{T \times 2d}$ and at the first compaction, $K_0$ and $V_0$ are raw chunks of KV, not compacted latents.

\paragraph{Continuous compaction implementations.}
Streaming H2O uses cumulative attention from the causal queries observed so far and forces a 64-token recent window. The StreamingLLM-style result protects the first four positions as attention sinks and re-rotates retained keys after eviction. $\delta$-mem writes its rank-8 state during ingestion and is frozen before the question, we test the state alone and with Recent-1024. LoCoCo uses
its pure learned convolutional merge with $M=1024$, 512-token chunks, and no
heavy-hitter branch. We train Qwen-specific per-layer convolution heads for 200 steps on RedPajama with the decoder frozen.



\section{Related Work}
\label{sec:related}

\paragraph{KV token retention.}
One KV compression strategy is to simply select and omit certain tokens from the cache, either as a function of the current context
\citep{zhang2023h2o,xiao2024efficient,li2024snapkv,cai2025pyramidkv},
or in a query-agnostic way
\citep{kim2025kvzip,chari2024compactor}. Learned variants predict future token
utility or distill a retention policy
\citep{moschella2026learning,chari2025kvdistill}. These methods are efficient,
but are limited in expressivity to the discrete token dictionary.

\paragraph{Synthetic KV representations.}
In contrast to token-side retention models, synthetic KV representations can generate examples in the full embedding space. Some methods optimize these KV vectors, either optimize a cache using self-study examples
\citep{eyuboglu2025cartridges} or reconstruct attention behavior with
closed-form solves \citep{zweiger2026fast}. Other methods train a model over a wider dataset to learn a KV compression model. LoCoCo merges cache states with learned
convolutions \citep{cai2024lococo}, while Still uses layer-wise latent queries
and cross-attention \citep{o2026still}. Our simplified Fast Diamond model belongs closest to this second
group, but studies why such writers are hard to train through a frozen model
and whether a simpler writer improves optimization.

\paragraph{Training models to use compacted KV representations.}
A number of methods also specifically train the base model to be able to read a token compression mechanism~\citep{mu2023gisting,chevalier2023adapting,ge2023incontextautoencoder, nawrot2024dynamic, li2026lclm}. While we likewise find in our paper that trainable weights helps for single-pass compression (compressing tokens a single time), it does not fully solve the recurrent case (iteratively compressing tokens).

\paragraph{Optimization through frozen models.}
Optimizing the inputs of a transformer or the KV cache has been explored in previous works~\citep{li2021prefix}. Prior work reports an unstable and sharp loss landscape~\citep{chen2023ptp}, and ameliorate this through input perturbations. Our work isolates specific properties and behaviors of the optimization landscape in the KV compaction regime.

\section{Discussion \& limitations}
Our simplified architecture performs well in our experiments, but the best way to add expressivity without sacrificing performance remains unclear. Methods that add expressivity, such as the methods we baseline against, don't perform as well as the heavily simplified architecture. However, this lunch does not come free, as when evaluated on iterated compaction regimes, the quality decays much faster than alternatives that can remain more expressive, although still the best performing method in the iterative context is well below the full-context threshold. Finding the right way to scale up expressivity of the KV compressor while allowing for repeated compaction remains an interesting direction for future work.

While our synthetic/linear environment and analysis provides more understanding for the phenomena often observed in KV compression, it is not exhaustive. A larger variety of small-parameter families and a series of increasingly-complex environments to analyze more complex phenomena both remain interesting directions for future analysis.

\section{Conclusion}
We studied the optimization problem created by learning compact KV caches through a frozen
  language model. Even a compression-free linear map displays important observed phenomena: a narrow basin around the global
  optimum, and broad, flat regions elsewhere. A number of fixes in the literature tie to these loss landscape attributes: denser losses, additional self-study
  queries, and weight relaxation, which help in restricted settings, but do not reliably carry over to
  compressive or recurrent caches.

  Motivated by the inflexibility of the optimization landscape, we show that a heavily-simplified Perceiver architecture outperforms baseline methods on MCQ over compressed caches, obtaining 92.5\% accuracy with 0.05 seconds of compaction per document. Iterative compaction remains a challenge for all methods, but in particular this simplified architecture. Finding the best balance of a compactor that is expressive enough to preserve long-term information while being able to reliably optimize through the difficult landscape of KV space remains an open and important problem for memory in continual learning systems.
\bibliographystyle{plainnat}
\bibliography{bibliography}

\begin{thebibliography}{24}
\providecommand{\natexlab}[1]{#1}
\providecommand{\url}[1]{\texttt{#1}}
\expandafter\ifx\csname urlstyle\endcsname\relax
  \providecommand{\doi}[1]{doi: #1}\else
  \providecommand{\doi}{doi: \begingroup \urlstyle{rm}\Url}\fi

\bibitem[Cai et~al.(2024)Cai, Tian, Wang, and Chen]{cai2024lococo}
Ruisi Cai, Yuandong Tian, Zhangyang Wang, and Beidi Chen.
\newblock {LoCoCo}: Dropping in convolutions for long context compression.
\newblock \emph{arXiv preprint arXiv:2406.05317}, 2024.
\newblock URL \url{https://arxiv.org/abs/2406.05317}.

\bibitem[Cai et~al.(2025)Cai, Zhang, Gao, Liu, Liu, Lu, Xiong, Dong, Hu, and Xiao]{cai2025pyramidkv}
Zefan Cai, Yichi Zhang, Bofei Gao, Yuliang Liu, Tianyu Liu, Keming Lu, Wayne Xiong, Yue Dong, Junyang Hu, and Wen Xiao.
\newblock {PyramidKV}: Dynamic {KV} cache compression based on pyramidal information funneling.
\newblock In \emph{Second Conference on Language Modeling}, 2025.
\newblock URL \url{https://openreview.net/forum?id=ayi7qezU87}.

\bibitem[Chari and Durme(2025)]{chari2024compactor}
Vivek Chari and Benjamin~Van Durme.
\newblock Compactor: Calibrated query-agnostic {KV} cache compression with approximate leverage scores.
\newblock \emph{arXiv preprint arXiv:2507.08143}, 2025.
\newblock URL \url{https://arxiv.org/abs/2507.08143}.

\bibitem[Chari et~al.(2025)Chari, Qin, and Durme]{chari2025kvdistill}
Vivek Chari, Guanghui Qin, and Benjamin~Van Durme.
\newblock {KV-Distill}: Nearly lossless learnable context compression for {LLMs}.
\newblock \emph{arXiv preprint arXiv:2503.10337}, 2025.
\newblock URL \url{https://arxiv.org/abs/2503.10337}.

\bibitem[Chen et~al.(2023)Chen, Chen, Huang, and Cheng]{chen2023ptp}
Lichang Chen, Jiuhai Chen, Heng Huang, and Minhao Cheng.
\newblock {PTP}: Boosting stability and performance of prompt tuning with perturbation-based regularizer.
\newblock In \emph{Proceedings of the 2023 Conference on Empirical Methods in Natural Language Processing}, pages 13512--13525, 2023.
\newblock \doi{10.18653/v1/2023.emnlp-main.833}.
\newblock URL \url{https://aclanthology.org/2023.emnlp-main.833/}.

\bibitem[Chevalier et~al.(2023)Chevalier, Wettig, Ajith, and Chen]{chevalier2023adapting}
Alexis Chevalier, Alexander Wettig, Anirudh Ajith, and Danqi Chen.
\newblock Adapting language models to compress contexts.
\newblock In \emph{Proceedings of the 2023 Conference on Empirical Methods in Natural Language Processing}, 2023.
\newblock URL \url{https://doi.org/10.18653/v1/2023.emnlp-main.232}.

\bibitem[Eyuboglu et~al.(2025)Eyuboglu, Ehrlich, Arora, Guha, Zinsley, Liu, Tennien, Rudra, Zou, Mirhoseini, and R{\'e}]{eyuboglu2025cartridges}
Sabri Eyuboglu, Ryan Ehrlich, Simran Arora, Neel Guha, Dylan Zinsley, Emily Liu, Will Tennien, Atri Rudra, James Zou, Azalia Mirhoseini, and Christopher R{\'e}.
\newblock Cartridges: Lightweight and general-purpose long context representations via self-study.
\newblock \emph{arXiv preprint arXiv:2506.06266}, 2025.
\newblock URL \url{https://arxiv.org/abs/2506.06266}.

\bibitem[Gao et~al.(2026)Gao, Wang, Jia, Zhang, Liu, Fan, and Xiao]{gao-etal-2026-sada}
Wenhao Gao, Tianlong Wang, Wei Jia, Linhao Zhang, Aiwei Liu, Miao Fan, and Zhou Xiao.
\newblock {SADA}: Bridging in-context learning and fine-tuning via state- aligned distillation adapters.
\newblock In \emph{Proceedings of the 64th Annual Meeting of the Association for Computational Linguistics (Volume 1: Long Papers)}, pages 22847--22862, San Diego, California, United States, July 2026. Association for Computational Linguistics.
\newblock URL \url{https://aclanthology.org/2026.acl-long.1046/}.

\bibitem[Ge et~al.(2024)Ge, Hu, Wang, Wang, Chen, and Wei]{ge2023incontextautoencoder}
Tao Ge, Jing Hu, Lei Wang, Xun Wang, Si-Qing Chen, and Furu Wei.
\newblock In-context autoencoder for context compression in a large language model.
\newblock In \emph{The Twelfth International Conference on Learning Representations}, 2024.
\newblock URL \url{https://arxiv.org/abs/2307.06945}.

\bibitem[Jiang and Jin(2026)]{jiang2026kvsculpt}
Bo~Jiang and Sian Jin.
\newblock {KVSculpt}: {KV} cache compression as distillation, 2026.
\newblock URL \url{https://arxiv.org/abs/2603.27819}.

\bibitem[Kim et~al.(2025)Kim, Kim, Kwon, Lee, Yun, and Song]{kim2025kvzip}
Jang-Hyun Kim, Jinuk Kim, Sangwoo Kwon, Jae~W. Lee, Sangdoo Yun, and Hyun~Oh Song.
\newblock {KVzip}: Query-agnostic {KV} cache compression with context reconstruction.
\newblock In \emph{Advances in Neural Information Processing Systems}, 2025.
\newblock URL \url{https://proceedings.neurips.cc/paper_files/paper/2025/hash/f4eaa4b8f2d08edb3f0af990d56134ea-Abstract-Conference.html}.

\bibitem[Lei et~al.(2026)Lei, Zhang, Li, Wang, Fan, Liu, Liu, Ma, Chen, and Poria]{lei2026deltamem}
Jingdi Lei, Di~Zhang, Junxian Li, Weida Wang, Kaixuan Fan, Xiang Liu, Qihan Liu, Xiaoteng Ma, Baian Chen, and Soujanya Poria.
\newblock {$\delta$-mem}: Efficient online memory for large language models.
\newblock \emph{arXiv preprint arXiv:2605.12357}, 2026.
\newblock URL \url{https://arxiv.org/abs/2605.12357}.

\bibitem[Li et~al.(2026)Li, McLeish, Chen, Kalra, Chen, Gazizov, Morisetty, Kailkhura, Menon, Liu, Bartoldson, Goldstein, Lotfi, Goldblum, and Izmailov]{li2026lclm}
Ang Li, Sean McLeish, Haozhe Chen, Nimit Kalra, Zaiqian Chen, Artem Gazizov, Venkata Anoop Suhas~Kumar Morisetty, Bhavya Kailkhura, Harshitha Menon, Zhuang Liu, Brian~R. Bartoldson, Tom Goldstein, Sanae Lotfi, Micah Goldblum, and Pavel Izmailov.
\newblock End-to-end context compression at scale.
\newblock \emph{arXiv preprint arXiv:2606.09659}, 2026.
\newblock URL \url{https://arxiv.org/abs/2606.09659}.

\bibitem[Li and Liang(2021)]{li2021prefix}
Xiang~Lisa Li and Percy Liang.
\newblock Prefix-tuning: Optimizing continuous prompts for generation.
\newblock In \emph{Proceedings of the 59th Annual Meeting of the Association for Computational Linguistics and the 11th International Joint Conference on Natural Language Processing (Volume 1: Long Papers)}, 2021.
\newblock URL \url{https://doi.org/10.18653/v1/2021.acl-long.353}.

\bibitem[Li et~al.(2024)Li, Huang, Yang, Venkitesh, Locatelli, Ye, Cai, Lewis, and Chen]{li2024snapkv}
Yuhong Li, Yingbing Huang, Bowen Yang, Bharat Venkitesh, Acyr Locatelli, Hanchen Ye, Tianle Cai, Patrick Lewis, and Deming Chen.
\newblock {SnapKV}: {LLM} knows what you are looking for before generation.
\newblock In \emph{Advances in Neural Information Processing Systems}, 2024.
\newblock URL \url{https://doi.org/10.52202/079017-0722}.

\bibitem[Moschella et~al.(2026)Moschella, Manduchi, and Sener]{moschella2026learning}
Luca Moschella, Laura Manduchi, and Ozan Sener.
\newblock Learning to evict from key-value cache.
\newblock In \emph{International Conference on Machine Learning}, 2026.
\newblock URL \url{https://arxiv.org/abs/2602.10238}.

\bibitem[Mu et~al.(2023)Mu, Li, and Goodman]{mu2023gisting}
Jesse Mu, Xiang~Lisa Li, and Noah Goodman.
\newblock Learning to compress prompts with gist tokens.
\newblock In \emph{Advances in Neural Information Processing Systems}, 2023.
\newblock URL \url{https://doi.org/10.52202/075280-0848}.

\bibitem[Nawrot et~al.(2024)Nawrot, {\L}a{\'n}cucki, Chochowski, Tarjan, and Ponti]{nawrot2024dynamic}
Piotr Nawrot, Adrian {\L}a{\'n}cucki, Marcin Chochowski, David Tarjan, and Edoardo~M. Ponti.
\newblock Dynamic memory compression: Retrofitting {LLMs} for accelerated inference.
\newblock In \emph{Proceedings of the 41st International Conference on Machine Learning}, pages 37396--37412, 2024.
\newblock URL \url{https://proceedings.mlr.press/v235/nawrot24a.html}.

\bibitem[O'Neill et~al.(2026)O'Neill, Sandomirsky, Partridge, Jayasekara, and Kirkby]{o2026still}
Charles O'Neill, Alex Sandomirsky, Harry Partridge, Mudith Jayasekara, and Max Kirkby.
\newblock Still: Amortized kv cache compaction in a single forward pass.
\newblock \emph{arXiv preprint arXiv:2606.07878}, 2026.
\newblock URL \url{https://arxiv.org/abs/2606.07878}.

\bibitem[{OneRec Team}(2026)]{onerec2026ksa}
{OneRec Team}.
\newblock Kwai summary attention technical report.
\newblock Technical report, Kuaishou Technology, 2026.
\newblock URL \url{https://arxiv.org/abs/2604.24432}.

\bibitem[Su et~al.(2024)Su, Murtadha, Lu, Pan, Wen, and Liu]{su2023roformerenhancedtransformerrotary}
Jianlin Su, Ahmed Murtadha, Yu~Lu, Shengfeng Pan, Bo~Wen, and Yunfeng Liu.
\newblock {RoFormer}: Enhanced transformer with rotary position embedding.
\newblock \emph{Neurocomputing}, 568:\penalty0 127063, 2024.
\newblock \doi{10.1016/j.neucom.2023.127063}.
\newblock URL \url{https://doi.org/10.1016/j.neucom.2023.127063}.

\bibitem[Xiao et~al.(2024)Xiao, Tian, Chen, Han, and Lewis]{xiao2024efficient}
Guangxuan Xiao, Yuandong Tian, Beidi Chen, Song Han, and Mike Lewis.
\newblock Efficient streaming language models with attention sinks.
\newblock In \emph{The Twelfth International Conference on Learning Representations}, 2024.
\newblock URL \url{https://arxiv.org/abs/2309.17453}.

\bibitem[Zhang et~al.(2023)Zhang, Sheng, Zhou, Chen, Zheng, Cai, Song, Tian, R{\'e}, Barrett, Wang, and Chen]{zhang2023h2o}
Zhenyu Zhang, Ying Sheng, Tianyi Zhou, Tianlong Chen, Lianmin Zheng, Ruisi Cai, Zhao Song, Yuandong Tian, Christopher R{\'e}, Clark Barrett, Zhangyang Wang, and Beidi Chen.
\newblock {H2O}: Heavy-hitter oracle for efficient generative inference of large language models.
\newblock In \emph{Advances in Neural Information Processing Systems}, 2023.
\newblock URL \url{https://doi.org/10.52202/075280-1506}.

\bibitem[Zweiger et~al.(2026)Zweiger, Fu, Guo, and Kim]{zweiger2026fast}
Adam Zweiger, Xinghong Fu, Han Guo, and Yoon Kim.
\newblock Fast {KV} compaction via attention matching.
\newblock \emph{arXiv preprint arXiv:2602.16284}, 2026.
\newblock URL \url{https://arxiv.org/abs/2602.16284}.

\end{thebibliography}

\clearpage

\clearpage
\appendix

\section*{Appendices}


\section{Detailed Methods}
\label{app:methods}

Initialization is block-diagonal identity-style. The cross-attention value projection and output projection are initialized as identity maps over the concatenated $[K;V]$ representation, and the frozen output heads extract the two halves:
\begin{equation}
    W_{\mathrm{key}} = [I_d \;\; 0], \qquad W_{\mathrm{value}} = [0 \;\; I_d].
\end{equation}
KVID initialization keeps the same identity K/V path but leaves the latent bank random and drops the diagonal query/key anchor.

Table~\ref{tab:hyperparameters} lists the canonical Fast Diamond training
recipe used for the Section~\ref{sec:full-method} results.

\begin{table}[H]
    \centering
    \caption{Canonical Fast Diamond hyperparameters.}
    \label{tab:hyperparameters}
    \small
    \begin{tabular}{ll}
        \toprule
        \textbf{Hyperparameter} & \textbf{Value} \\
        \midrule
        \multicolumn{2}{l}{\emph{Model}} \\
        Base model & Qwen3-4B (frozen) \\
        Precision & BF16 \\
        Attention & FlashAttention-2 \\
        \midrule
        \multicolumn{2}{l}{\emph{Architecture}} \\
        Perceiver blocks & 1 (no self-attention, no residual) \\
        Cross-attention heads & 1 \\
        Latent dimension & 256 \\
        Latent queries & direct (no learned projection) \\
        K/V output projections & frozen \\
        Latent initialization & block-diagonal identity \\
        RoPE base & 10 \\
        Attention logit scaling & $\mathrm{softmax}(d_\ell\, \tilde q \tilde k^\top)\, \tilde v$ \\
        \midrule
        \multicolumn{2}{l}{\emph{Data}} \\
        Input context & 8,192 tokens \\
        Compact cache & 1,024 latents \\
        Domains & Financial, Gutenberg, Code, Legal \\
        Cue tokens & 64 \\
        Max continuation tokens & 128 \\
        \midrule
        \multicolumn{2}{l}{\emph{Optimizer}} \\
        Optimizer & AdamW \\
        Learning rate & $3 \times 10^{-4}$ \\
        $\beta_1, \beta_2$ & $0.9, 0.999$ \\
        LR schedule & cosine \\
        Cosine floor & $0.1 \times$ peak LR \\
        Warmup steps & 100 \\
        Training steps & 3,000 \\
        \midrule
        \multicolumn{2}{l}{\emph{Batching}} \\
        Micro-batch size & 8 \\
        Gradient accumulation & 1 \\
        GPUs & $8 \times$ H200 \\
        Effective batch size & 64 \\
        \midrule
        \multicolumn{2}{l}{\emph{Objective}} \\
        Loss & forward KL on answer tokens \\
        KL normalization & per-example no-context KL \\
        Normalization $\varepsilon$ & $10^{-6}$ \\
        \bottomrule
    \end{tabular}
\end{table}

\section{Compute resources}
\label{app:compute-resources}

All GPU-backed training and evaluation jobs were run on Baseten cloud workers. Across the successful paper-facing runs underlying the main-text and appendix results, we estimate approximately 2,800–3,300 NVIDIA H100 GPU-hours and 800–1,000 NVIDIA H200 GPU-hours. Most full-scale Qwen3-4B experiments used eight-GPU nodes; the toy self-study query-count experiment used four to eight H100 GPUs. These estimates count each underlying training or evaluation lineage once when multiple figures reuse the same checkpoint. Preliminary, failed, and diagnostic jobs required
additional exploratory compute beyond these paper-facing runs.


\section{Attention MSE}
\label{app:a-mse}

The compact readout cross-attends $n<<N$ latent queries
$Q \in \mathbb{R}^{n \times 2d}$ against cache keys and values
$K, V \in \mathbb{R}^{N \times 2d}$. Because the attention output lives
in query space,
\begin{equation}
    O = \mathrm{softmax}(Q K^\top) V \in \mathbb{R}^{n \times 2d},
\end{equation}
the compact cache and the full cache produce outputs of the same size, and
the attention-MSE objective is computable between any cache sizes:
\begin{equation}
    \mathcal{L}_{\mathrm{attn}} = \lVert O - O_{\mathrm{full}} \rVert^2,
    \qquad
    O_{\mathrm{full}} = \mathrm{softmax}(Q K_{\mathrm{full}}^\top)
    V_{\mathrm{full}}.
\end{equation}
The softmax can use either the student's own queries or the teacher's,
and the readout produces compact keys and values
$C_k, C_v \in \mathbb{R}^{n \times 2d}$. We
compare three query modes at $1k\to128$ (8x) compression under one 300-update
budget: student queries for both (SS), teacher queries for teacher
with student queries for the student (TS), and
teacher queries for both (TT). 

\begin{figure}[H]
    \centering
    \includegraphics[width=0.95\linewidth]{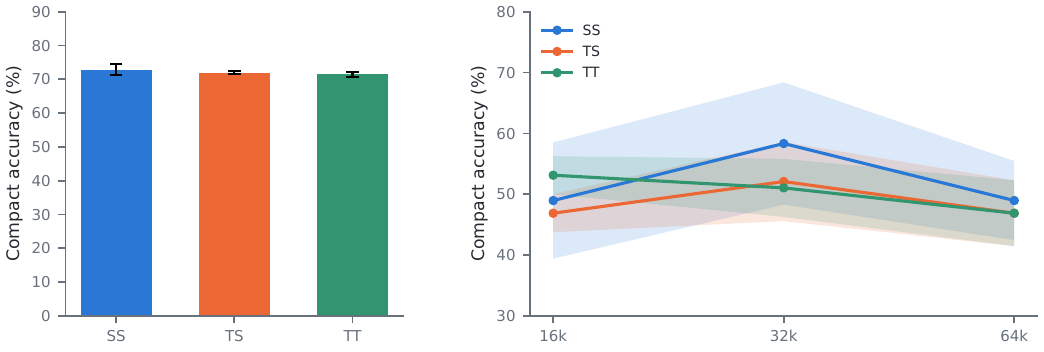}
    \caption{Query-source ablation for attention MSE (with normalized KL)
    at 1k-to-128 (300 updates). Left: final 1k compact accuracy on 184
    questions
    Right: compact accuracy at 16k/32k/64k in the iterative compaction setting. The differences between
    query modes is small at every horizon. SS: student queries, TS: teacher queries for teacher and student queries for the student, and TT: teacher queries for both.}
    \label{fig:app-qsource}
\end{figure}

Figure~\ref{fig:app-qsource} shows that the choice matters
little: the three arms land within 2.2 points of each other on the final 1k
evaluation, so SS is selected.

\begin{figure}[H]
    \centering
    \includegraphics[width=0.58\linewidth]{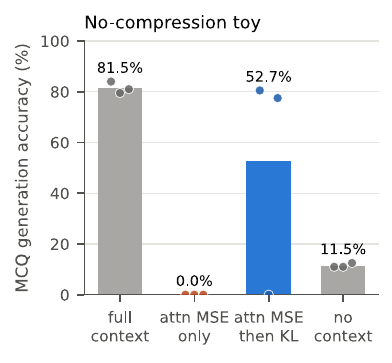}
    \caption{Attention MSE is not a behavioral certificate in the
    no-compression toy. Training on attention MSE alone reaches proxy loss
    $0.021$ yet answers $0/200$ questions, while attention MSE followed by a
    KL phase can reach $80.5\%$.}
    \label{fig:app-attnmse-counterexample-toy}
\end{figure}

\begin{figure}[H]
    \centering
    \includegraphics[width=0.98\linewidth]{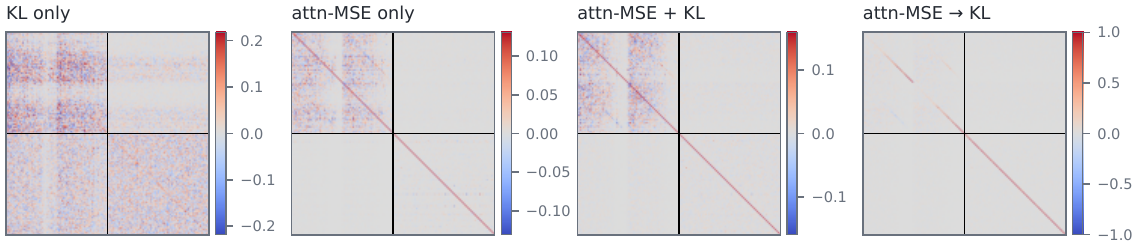}
    \caption{Loss isolation with a single shared $256{\times}256$
    token-linear compactor matrix (no perceiver, no compression).}
    \label{fig:app-single-matrix}
\end{figure}

Section~\ref{sec:landscape-fixes} reports that pre-conditioning with
attention MSE before applying KL is the strongest recipe in the toy setting.
Figure~\ref{fig:app-attnmse-counterexample-toy} shows that attention MSE alone
does not certify behavior in the no-compression toy.
Figure~\ref{fig:app-single-matrix} isolates the loss from the architecture
with a single shared weight matrix: KL alone never forms an identity
diagonal, and although attention-MSE does, the resulting cache answers 0/200
held-out questions. Pre-conditioning
with attention MSE first reaches an identity-like state (identity cosine
0.838 at the handoff) and KL finetuning from such a state saturates 
utility at no compression.

\begin{figure}[H]
    \centering
    \includegraphics[width=0.95\linewidth]{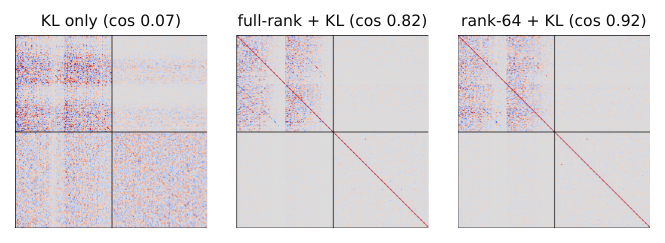}
    \caption{Terminal $256 \times 256$ compactor weight matrices for KL-only, full-rank attention MSE + KL, and rank-64 attention MSE + KL. The $2 \times 2$ block structure reflects the concatenated
    $[K; V]$ representation.}
    \label{fig:toy-lowrank-matrices}
\end{figure}

We ablate low-rank projections of $K$ and $Q$ prior to attention MSE calculation ($K$'s basis is determined via SVD decomposition and applied to both $K$ and $Q$). Figure~\ref{fig:toy-lowrank-matrices} shows the terminal weight matrices. KL-only produces unstructured weights but the attn-MSE + KL arms recover more block-diagonal structure. Low-rank projections of $K$ and $Q$ have neutral to positive effect on identity discovery.

\section{Latent query ablation}
\label{app:z-transfer}

The toy still setting outlined a key deficiency in the offline training of Still, no inter-layer communication between perceivers. Errors cannot be progressively corrected from layer to layer in real-time. We ablate training the linear map as a global map, as well as a layer, head, and per layer and per head map to identify if this lack of communication is a bottleneck, and if training Z is a necessity at all. We find that at least in this formulation, training Z is beneficial, and constraining Z to a global weight is neutral-to-positive.

Firstly, by copying an attention MSE  learned global Toy matrix into layer, head, and
per layer per head parameterizations we can better restrict the linear map to the global basin before applying KL.

\begin{figure}[H]
    \centering
    \includegraphics[width=0.98\linewidth]{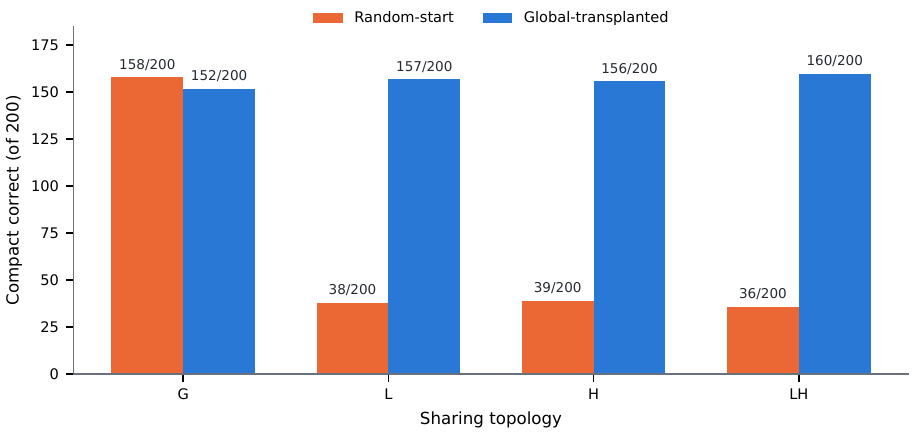}
    \caption{Toy Still training starting from a random linear map before 1,000 attention-MSE and 500 KL updates. Copying the learned global matrix into each topology before KL finetuning greatly improves performance. Globally training Z helps to find a useful initialization.}
    \label{fig:app-toy-z-basin-transfer}
\end{figure}

Secondly, using the global training result to the Still setting, we remark that the larger share of the splitting gain is attributable to K/V/O sharing capacity. Here K/V/O refer to the $W_K$, $W_V$, and $W_O$ projection matrices in the Fast Diamond architecture.

\begin{figure}[H]
    \centering
    \includegraphics[width=0.98\linewidth]{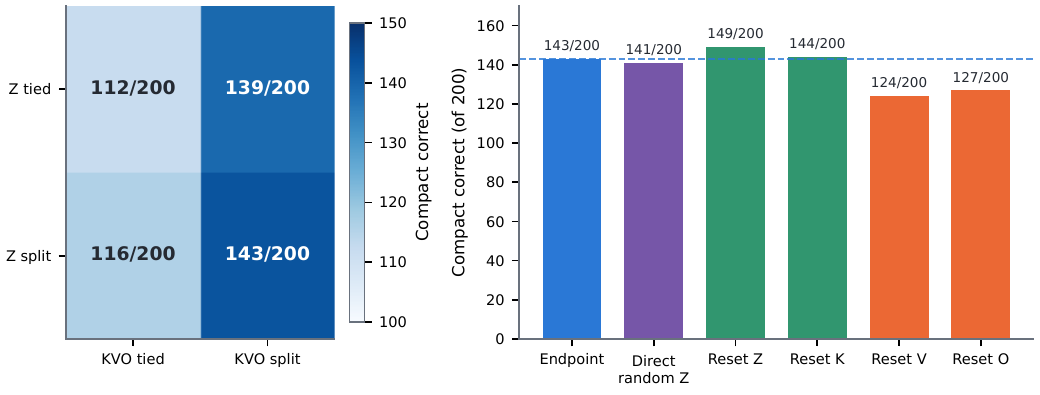}
    \caption{Global-Z/KVO sharing factorial and checkpoint-reset diagnostics in
    full Still. Splitting K/V/O adds $27$ correct answers at either Z topology, whereas splitting Z only adds $4$. Resetting refers to re-randomizing the weight prior to the KL training phase. Resetting V or O slightly hurts performance and all other resets are neutral. Counts are compact-correct items, $n=200$.}
    \label{fig:global-z-kvo-mechanism}
\end{figure}

Finally, we ablate restricting the optimization to only the $W_K$, $W_V$, and $W_O$ matrices, freezing Z to its pseudo-diagonal initialization across all layers. The effect is a greater than 10x reduction in parameters and a loss of close to 10 points of utility over a 1000 step training run. Training Z raises the performance ceiling of our compactor, but trades much more parameter space per unit of utility than our projection matrices.

\begin{figure}[H]
    \centering
    \includegraphics[width=0.72\linewidth]{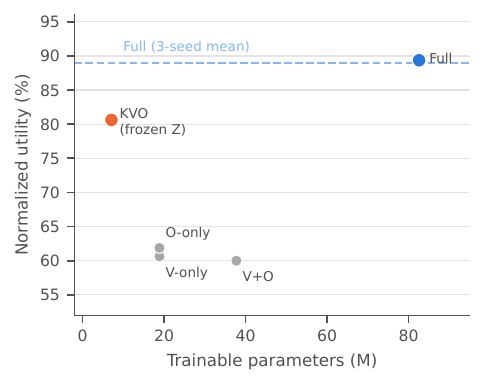}
    \caption{Normalized utility versus trainable Still parameters ($1024\to128$). Training "KVO" ($W_K$, $W_V$, $W_O$ matrices) with frozen Z uses $11.65\times$ fewer parameters than
    Full and only loses $8.75$ utility points. V-only, O-only, and V+O cluster
    near $60\%$ regardless of parameter count.}
    \label{fig:app-parameter-reduction}
\end{figure}



\section{Local landscape around the historical plateau}
\label{app:local-zoom}

\begin{figure}[H]
    \centering
    \includegraphics[width=0.55\linewidth]{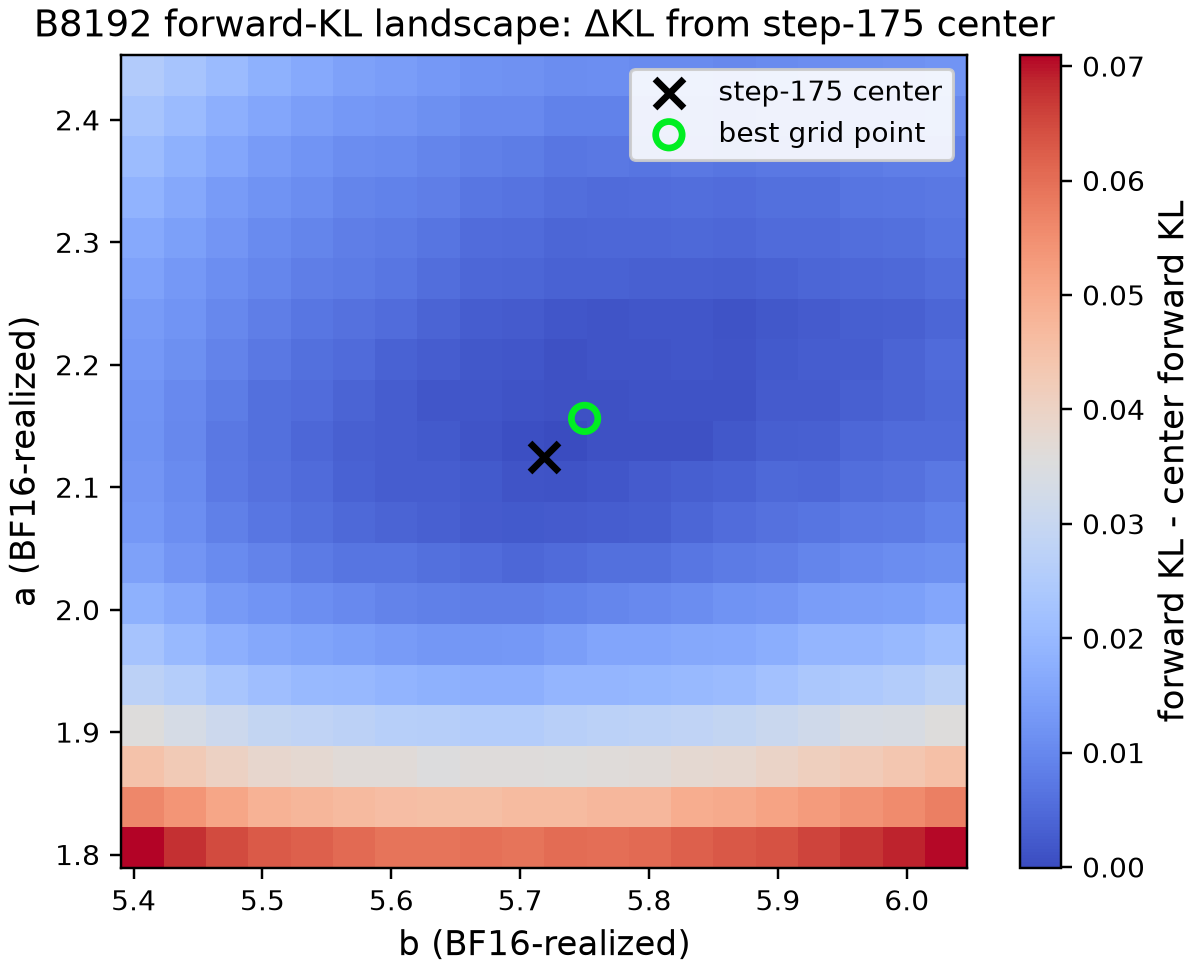}
    \caption{Local $20\times20$ sweep of $20\times20$ distinct
    BF16-realizable values around the step-175 plateau point
    $(2.1206,5.7100)$, on the same cohort and color scale convention as
    Figure~\ref{fig:kvmix2x2-landscape}. The plateau is flat at BF16
    resolution.}
    \label{fig:kvmix2x2-landscape-local}
\end{figure}

\section{Weight-relaxation ablation}
\label{app:w-ablate}

Section~\ref{sec:fix-relaxing} describes weight relaxation in prose;
this appendix gives the concrete objective: each relaxed weight is defined as $W_j = W_j^{(0)} + \Delta W_j$. The return
objective at update $u$ is:
\begin{equation}
    \mathcal{L}_u = (1-\alpha_u)\mathcal{L}_{\mathrm{KL}}
    + \alpha_u \sum_j \lVert \Delta W_j \rVert_F^2,
    \qquad \alpha_u = \frac{u-1}{U-1}.
    \label{eq:weight-relaxation}
\end{equation}
The penalty coefficient anneals linearly from 0 to 1 over the $U$ updates,
so training moves from pure task KL to a pure pull back to the frozen
weights.

\begin{figure}[H]
    \centering
    \includegraphics[width=0.7\linewidth]{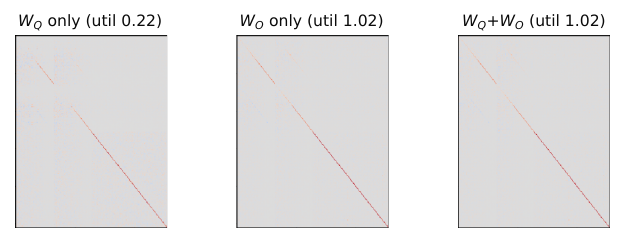}
    \caption{Terminal compactor weight matrices for $W_Q$-only,
    $W_O$-only, and $W_Q + W_O$. All begin to approach identity
    however only arms training $W_O$ saturate task utility. The setting is the no-compression toy trained for 6,000 updates.}
    \label{fig:toy-wo-scope-matrices}
\end{figure}

In the toy setting weight relaxation helps KL-only global linear maps to find identity like structure, requiring as little as only $W_O$ to be relaxed to gain the benefit. Figure~\ref{fig:toy-wo-scope-matrices} shows the terminal compactor weight
matrices for linear maps in toy still trained under the weight-relaxation objective. 

\begin{figure}[H]
    \centering
    \includegraphics[width=0.6\linewidth]{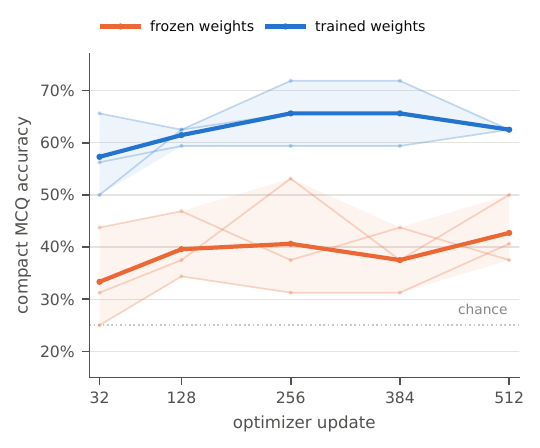}
    \caption{Weight training is necessary for recurrent application of the linear map with KL loss. Horizontal bars mark the family median.
    Weight-trained arms cluster near $60$--$70\%$; frozen-weight arms
    stay at or below $40\%$.}
    \label{fig:recurrent-weight-training}
\end{figure}

In the recurrent toy setting the linear map performed poorly under recurrence, neither having strong identity structure or high utility. Weight training being the strongest intervention found, we ablate its inclusion in the recurrent recipe. Figure~\ref{fig:recurrent-weight-training} shows that under recurrence this intervention seems not only helpful, but necessary---unfortunately, weight relaxation does not recover the same benefit.

\section{Architectural ablation}
\label{app:a-ablate}

In pursuit of increasing compact performance of Still, we ablate changes in both cross-attention RoPE and the perceiver formulation. We find that the inclusion of RoPE consistently improves performance and that the canonical cross-attention formulation is optimal against all ablated changes.

\begin{figure}[H]
    \centering
    \includegraphics[width=0.98\linewidth]{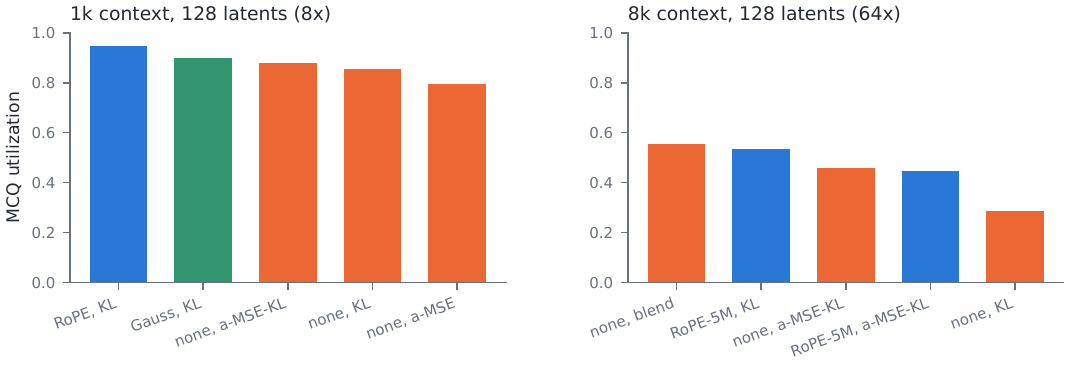}
    \caption{Single-pass compaction endpoints with different cross-attention
    position encodings. Left: 1k-to-128 (8$\times$ compression). Right: 8k-to-128 (64$\times$). RoPE consistently helps performance and attention MSE is neutral to negative for utility.}
    \label{fig:app-rope-ladder}
\end{figure}

Figure~\ref{fig:app-rope-ladder} compares stills with matched training recipes varying only in the application of RoPE prior to the cross-attention. These runs are in the toy still setting and leverage KL and attention-MSE losses. the KL-only pair differs modestly at 8$\times$ (0.947 vs.\ 0.853) and
separates widely at 64$\times$ (0.536 vs.\ 0.288), with attention-MSE
variants between the poles.

\begin{figure}[H]
    \centering
    \includegraphics[width=0.98\linewidth]{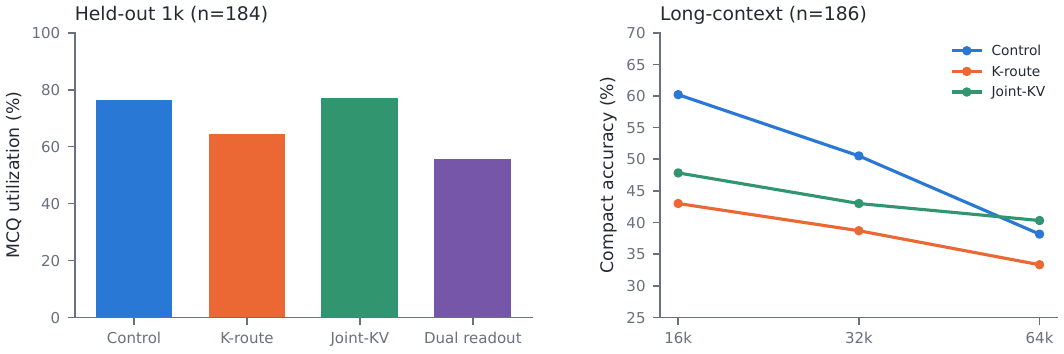}
    \caption{Routing and readout ablation, single-pass 1k-to-128 over
    301 training steps, canonical KL-loss. Left: Terminal held-out utilization at still training horizon. Right Long-context compact accuracy at
    16k/32k/64k in the iterative compaction setting. Control outperforms in all settings with only Joint-KV recovering similar performance in the 8k and 64k settings.}
    \label{fig:app-routing-readout}
\end{figure}

Still builds compact K/V from one joint perceiver state: the perceiver
maps the cache to latent queries $Z$, and the input is mapped as
$\tilde k = X W_K$, $\tilde v = XW_V$. We compare three
alternatives. K-route uses a smaller $Z$ of dim equal to $K$: $Z \in \mathbb{R}^{n \times d}$ and computes one routing map from the keys and applies it
to both: $A = \mathrm{softmax}(Z \tilde k^\top)$, $C_k = A K$, $C_v = A V$.
Joint-KV computes the map as in the canonical formulation and then applies them to $K$ and $V$ independently after down projection:
$A = \mathrm{softmax}(Z [K;V]^\top)$, $C_k = W_K A K$, $C_v = W_V A V$. Direct
dual readout gives K and V independent attentions:
$C_k = \mathrm{Attn}(Z, K, K)$,
$C_v = \mathrm{Attn}(Z, V, V)$.
Figure~\ref{fig:app-routing-readout} compares all four.

\begin{figure}[H]
    \centering
    \includegraphics[width=0.98\linewidth]{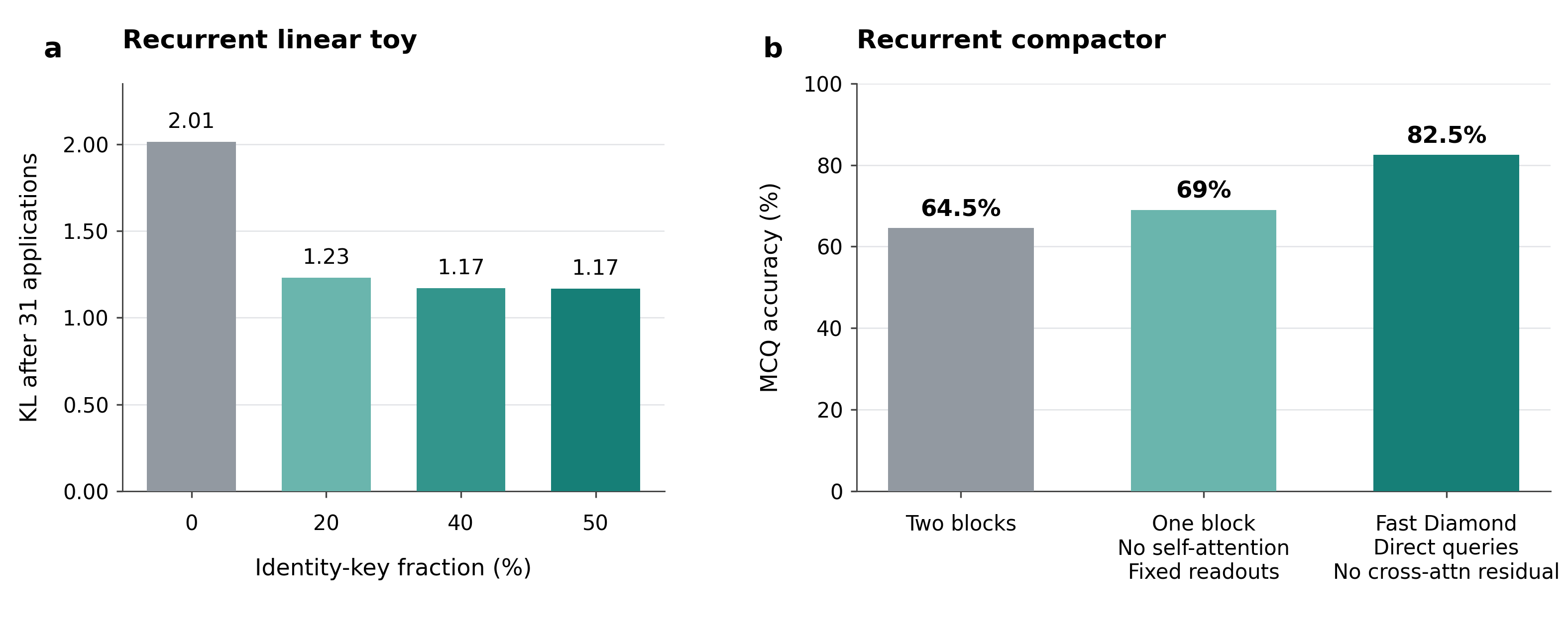}
    \caption{Left (a): Recurrent linear-map KL after 31 applications, following training with seven applications. Identity-key fractions of 0\%, 20\%, 40\% and 50\% give KL values of 2.01, 1.23, 1.17 and 1.17. Right (b): Recurrent architecture simplification. Removing the second block and self-attention and fixing the output readouts improves accuracy; direct queries and removal of the cross-attention residual give a further gain. Results use the same 200-question evaluation.}
    \label{fig:transition}
\end{figure}

Figure~\ref{fig:transition} details the gains seen in recurrent compaction due to freezing and simplifying architectures. Architecture simplification sees accuracy rise from 64.5\% to 69.0\% to 82.5\%, while KL falls from 0.420 to 0.396 to 0.251. Motivated by the increase in frozen keys in recurrent toy still compaction lowering KL.

\raggedbottom
\section{Higher compression results}

We provide detailed evaluation results across all methods in this section. We additionally evaluate the weight-trained Fast Diamond--KVID variant which achieves utility at compression ratios of: $89.49\%$ at $8\times$, $75.58\%$ at $16\times$, and $56.26\%$ at $64\times$. This result indicates that with base model weight training the more restrictive block-diagonal identity-initialization is not strictly necessary, though it still routinely outperforms the KVID initialization in most settings (all but 64x compression).

\begin{figure}[!htbp]
\centering
\includegraphics[width=\textwidth]{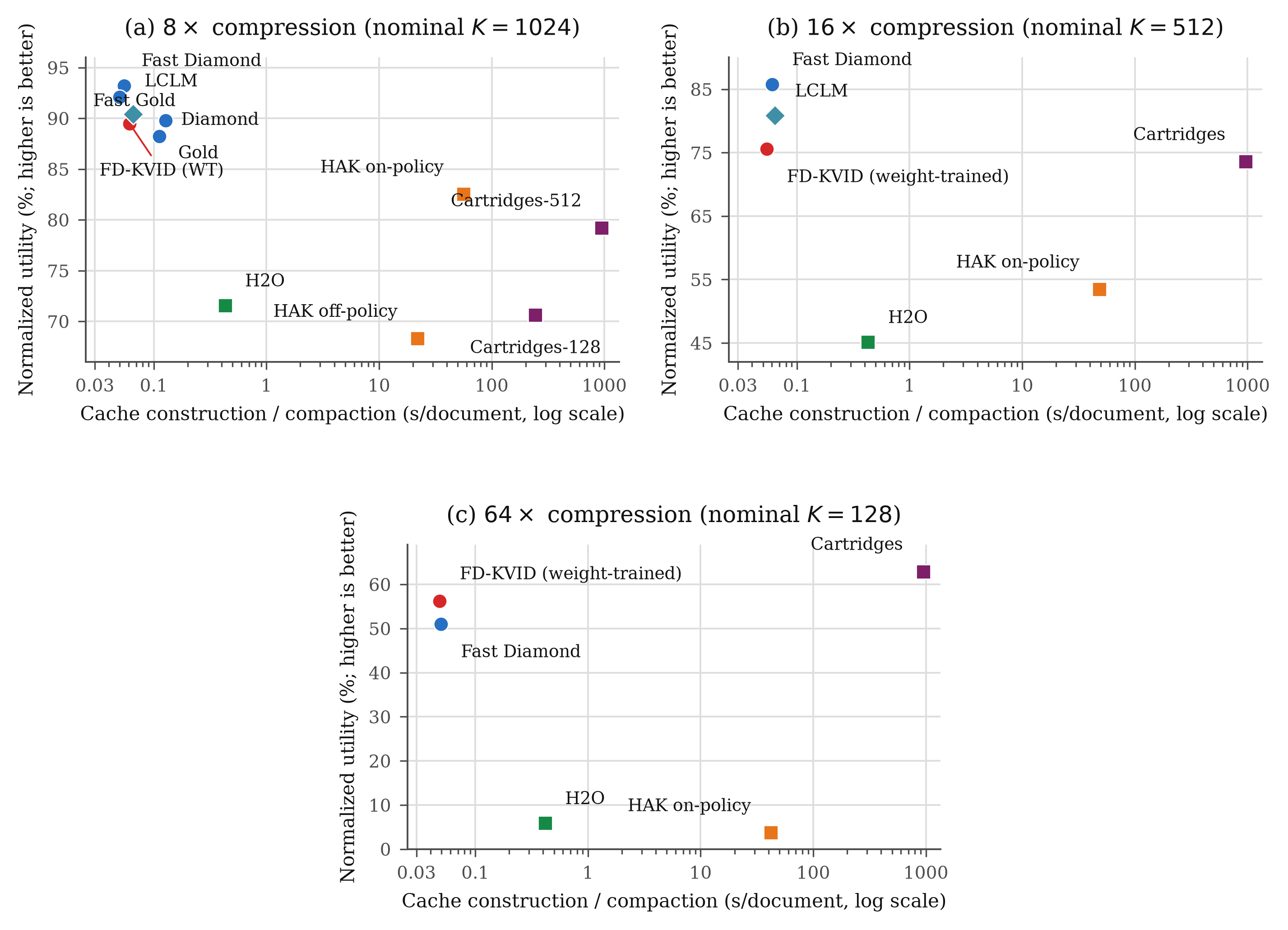}
\caption{Speed--utility comparison for the compressed-cache methods.
The horizontal axis is logarithmic. 
Red FD--KVID points are weight-trained Fast Diamond variants evaluated with their coupled trained base model weights; blue Fast Diamond points use the standard base model.}
\label{fig:matched-speed-utility-full}
\end{figure}


\section{Full comparative results}

\begin{table}[H]
\centering
\caption{Main results on MCQ compaction. H2O, HAK, and Still measure online compaction time, LCLM reports encoder-plus-adapter compaction time, and Cartridges reports offline synthesis plus per-document optimization time.}
\label{tab:matched-baseline-results-full}
\small
\resizebox{\textwidth}{!}{%
\begin{tabular}{lrrrrr}
\toprule
Method &
Accuracy &
Normalized utility &
Cache / compaction &
Generation \\
&
{(\%)} &
{(\%)} &
{(s/document)} &
{(s/question)} \\
\midrule
Full-context control
    &
    98.29 & 100.00 & 0.00 & 8.44 \\
Fast Diamond (one block)
    &
    92.48 & 93.20 & 0.05 & 10.89 \\
Fast Gold
    &
    91.56 & 92.12 & 0.05 & 10.05 \\
LCLM mean-pooling 8$\times$\textsuperscript{$\ast$}
    &
    90.11 & 90.42 & 0.07 & 1.45 \\
Fast Diamond--KVID (weight-trained)
    &
    89.31 & 89.49 & 0.06 & 11.88 \\
Diamond (two blocks)
    &
    89.58 & 89.80 & 0.13 & 12.23 \\
Gold (Still)
    &
    88.26 & 88.25 & 0.11 & 10.08 \\
HAK on-policy
    &
    83.38 & 82.53 & 56.26 & 16.18 \\
Cartridges--512 (long)\textsuperscript{$\dagger$}
    &
    80.61 & 79.20 & 947.19 & 9.65 \\
H2O
    &
    74.01 & 71.56 & 0.43 & 10.61 \\
Cartridges--128 (short)
    &
    73.22 & 70.64 & 244.16 & 11.41 \\
HAK off-policy
    &
    71.24 & 68.32 & 21.88 & 17.37 \\
$\delta$-mem + recent-1024 KV
    &
    25.59 & 14.84 & 0.42 & 3.11 \\
Recent-1024 KV only
    &
    22.43 & 11.13 & 0.24 & 1.84 \\
$\delta$-mem state only
    &
    12.66 & -0.31 & 0.64 & 3.53 \\
No-context control
    &
    12.93 & 0.00 & 0.00 & 10.13 \\
\bottomrule
\end{tabular}%
}

\vspace{2pt}
\end{table}

\section{More landscape}

\begin{figure}[H]
    \centering
    \begin{subfigure}[t]{0.425\linewidth}
        \centering
        \includegraphics[width=\linewidth]{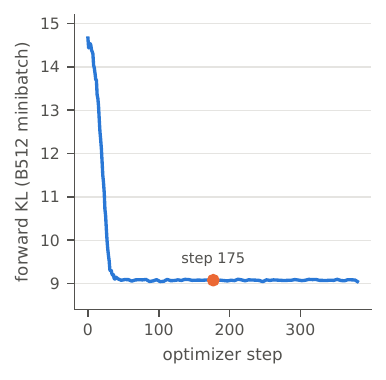}
        \caption{Training from $(a,b)=(3,3)$: the B512 stochastic forward
        KL falls from $14.66$ to about $9.13$ in the first 40 updates,
        then stays on the same plateau through step 175.}
        \label{fig:kvmix2x2-training}
    \end{subfigure}\hfill
    \begin{subfigure}[t]{0.53\linewidth}
        \centering
        \includegraphics[width=\linewidth]{figures/fig_kvmix2x2_landscape_broad.png}
        \caption{Broad $22\times33$ direct-evaluation sweep containing
        identity at $(1,0)$ and the step-175 point near $(2.12,5.71)$.}
        \label{fig:kvmix2x2-landscape-broad-appendix}
    \end{subfigure}
    \caption{Applying the KV linear map on a single single-layer/head run of the
    two-parameter transform, and the loss surface it sits on. Every surface
    point uses the same 8,192 examples and 745,203 answer tokens; colors show
    the answer-token-weighted forward-KL difference from the step-175 center.
    These are direct evaluations, not optimizer updates. A finer local sweep
    around the plateau point is in Appendix~\ref{app:local-zoom}.}
    \label{fig:kvmix2x2-landscape}
\end{figure}

\end{document}